\documentclass[conference]{IEEEtran}
\usepackage{cite}
\usepackage{flushend}
\usepackage[pdftex]{graphicx}
\usepackage{makecell}
\usepackage{multirow}
\usepackage{booktabs}
\usepackage{pifont}
\usepackage{xcolor,colortbl}
\usepackage{hyperref}
\begin{document}

\title{CyberMetric: A Curated Benchmark Dataset for Evaluating Cybersecurity Knowledge of LLMs using RAG with Human Validation}
\title{CyberMetric: Creating a Benchmark Dataset RAG with Human Validation for Evaluating Cybersecurity Knowledge of LLMs}

\title{CyberMetric: A Benchmark Dataset based on Retrieval-Augmented
Generation for Evaluating LLMs in Cybersecurity Knowledge}

\author{
\IEEEauthorblockN{Norbert Tihanyi, Mohamed Amine Ferrag, Ridhi Jain}
\IEEEauthorblockA{Technology Innovation Institute (TII)\\
Abu Dhabi, United Arab Emirates\\}
\and
\IEEEauthorblockN{Tamas Bisztray}
\IEEEauthorblockA{University of Oslo\\
Oslo, Norway}
\and
\IEEEauthorblockN{Merouane Debbah}
\IEEEauthorblockA{Khalifa University\\
Abu Dhabi, United Arab Emirates
}
}


%


\maketitle

\begin{abstract}
Large Language Models (LLMs) are increasingly used across various domains, from software development to cyber threat intelligence. Understanding all the different fields of cybersecurity, which includes topics such as cryptography, reverse engineering, and risk assessment, poses a challenge even for human experts. To accurately test the general knowledge of LLMs in cybersecurity, the research community needs a diverse, accurate, and up-to-date dataset. To address this gap, we present CyberMetric-80, CyberMetric-500, CyberMetric-2000, and CyberMetric-10000, which are multiple-choice Q\&A benchmark datasets comprising 80, 500, 2000, and 10,000 questions respectively. By utilizing GPT-3.5 and Retrieval-Augmented Generation (RAG), we collected documents, including NIST standards, research papers, publicly accessible books, RFCs, and other publications in the cybersecurity domain, to generate questions, each with four possible answers. The results underwent several rounds of error checking and refinement. Human experts invested over 200 hours validating the questions and solutions to ensure their accuracy and relevance, and to filter out any questions unrelated to cybersecurity. We have evaluated and compared 25 state-of-the-art LLM models on the CyberMetric datasets. In addition to our primary goal of evaluating LLMs, we involved 30 human participants to solve CyberMetric-80 in a closed-book scenario. The results can serve as a reference for comparing the general cybersecurity knowledge of humans and LLMs.
The findings revealed that GPT-4o, GPT-4-turbo, Mixtral-8x7B-Instruct, Falcon-180B-Chat, and GEMINI-pro 1.0 were the best-performing LLMs. Additionally, the top LLMs were more accurate than humans on CyberMetric-80, although highly experienced human experts still outperformed small models such as Llama-3-8B, Phi-2 or Gemma-7b. The CyberMetric dataset is publicly available for the research community and can be downloaded from the projects' website: \url{https://github.com/CyberMetric}.
\end{abstract}


%
\IEEEpeerreviewmaketitle

\section{Introduction}

The Industrial Revolution in the 18th century initiated a technological era, marked by significant advancements such as the steam engine, which exceeded the efficiency of human and animal labor. The mid-20th century saw the UNIVAC~\cite{UNIVAC} computer in the 1950s, performing complex calculations and data processing faster than any human. By the 1970s, early chess programs began challenging experienced human players, demonstrating the evolving potential of Artificial Intelligence (AI). In 1997, IBM's Deep Blue~\cite{deepblue} defeated Garry Kasparov~\cite{hsu_behind_2022}, marking a groundbreaking moment in AI. This trend continued in 2016 with AlphaGo~\cite{alphago}, developed by Google DeepMind, outclassing world champion Lee Sedol in Go~\cite{chouard_go_2016}.

Large Language Models (LLMs) have revolutionized Natural Language Processing (NLP), improving automated text generation and enabling human-like interactions. These breakthroughs extend into sectors like medicine~\cite{singhal_large_2023}, finance~\cite{wu_bloomberggpt_2023}, and notably, cybersecurity~\cite{ferrag_securefalcon_2023}. LLMs offer immense potential in cybersecurity, enhancing domains from threat detection~\cite{ferrag_revolutionizing_2023} to policy interpretation \cite{policy}.

Cybersecurity encompasses diverse topics that can require a strong mathematical foundation, programming, creative thinking, analytical skills, and managerial tasks, making it a long and challenging process for humans to master. The two research questions naturally rises:

\begin{itemize}
  \item \textbf{RQ1:} Has machine intelligence already surpassed humans in answering questions across the entire breadth of cybersecurity knowledge in a closed-book test?
 \item \textbf{RQ2:} Which currently available model achieves the highest accuracy in answering questions across diverse cybersecurity domains?

\end{itemize}

To address these questions, two prerequisites must be met: \textit{1)} creating a trustworthy and validated large dataset to assess LLMs' cybersecurity knowledge; \textit{2)} selecting a subset of questions to match humans against LLMs. Although there are existing datasets and surveys focused on problem solving, coding, penetration testing, or threat intelligence~\cite{zhang2024llms,hou2023large}, a comprehensive dataset for testing broad cybersecurity knowledge is still lacking. This paper aims to bridge this gap by developing \texttt{CyberMetric}, the first comprehensive benchmark dataset for evaluating LLMs' expertise across the field of cybersecurity. Our dataset includes 9 domains: Disaster Recovery and BCP, Identity and Access Management (IAM), IoT Security, Cryptography, Wireless Security, Network Security, Cloud Security, Penetration Testing, and Compliance/Audit. The dataset comprises 10,000 questions and answers, extracted from hundreds of guidelines, standards, books, and research papers, totalling over 100,000 pages. 
The main contributions of this paper can be summarized as follows:
\begin{enumerate}

\item We present \textit{CyberMetric-10000} a comprehensive benchmark dataset that includes 10,000 cybersecurity-related questions designed to evaluate the understanding of LLMs across nine distinct domains within cybersecurity.
In addition, we created smaller subsets of the original datasets, named \textit{CyberMetric-80}, \textit{CyberMetric-500}, and \textit{CyberMetric-2000}. The 80 and 500 datasets are fully validated by human experts. 
\textit{CyberMetric} has been made accessible to the research community as a foundational metric for cybersecurity knowledge at \url{https://github.com/CyberMetric}.

\item We conducted an extensive user study with 30 human experts answering \textit{CyberMetric-80} to match their performance against LLMs. This analysis aims to understand how human expertise compares to machine intelligence in terms of cybersecurity knowledge when it comes to answering multiple-choice questions.

\end{enumerate}

The rest of the paper is organized as follows: Section~\ref{sec:related} overviews related literature, Section~\ref{sec:methodology} details the methodology and dataset creation, Section~\ref{sec:experiments} discusses the experimental setup, objectives, and results, Section~\ref{sec:discussion} presents our observations, Section~\ref{sec:limit} discusses limitations and ethical considerations, and Section~\ref{sec:conclusion} concludes the paper.


\begin{figure*}[ht] 
\centering
\includegraphics[width=1\textwidth]{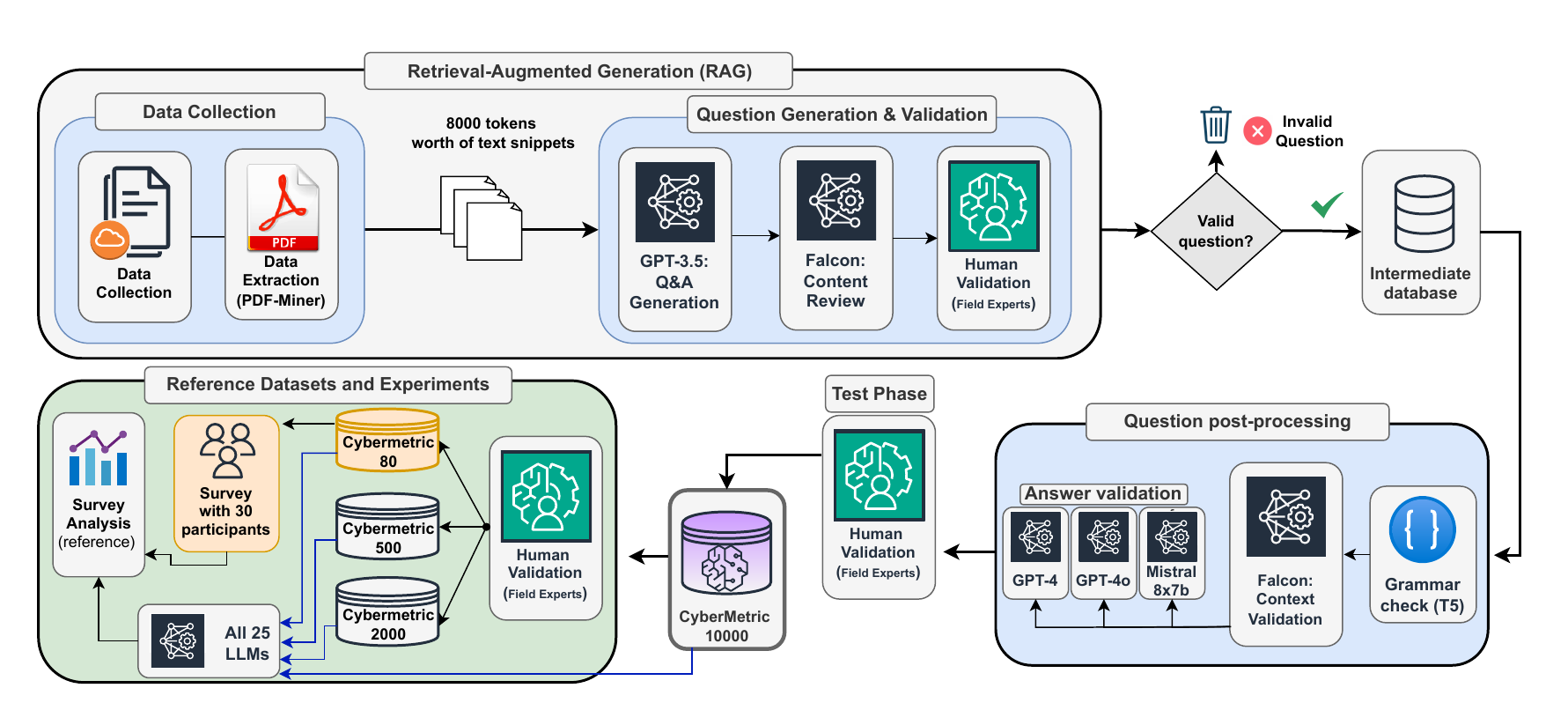}
\caption{Framework for AI-driven question generation methodology, incorporating human validation.}
\label{fig:Framework}
\end{figure*}

\section{Related work}
\label{sec:related}
As LLMs have evolved, the need for domain-specific benchmark datasets to test and compare their capabilities has grown significantly. These datasets are crucial for assessing LLMs in different cybersecurity domains, guiding further development and training efforts. While datasets exists for mathematical problem-solving, coding, and reasoning~\cite{chen2021codex,hendrycksmath2021,dibia-etal-2023-aligning}, large-scale datasets for assessing broad cybersecurity knowledge have not been developed prior to our research.
Curating Q\&A datasets has gained popularity in recent years. Khot et al.~\cite{khot2020qasc} introduced QASC, a multiple-choice question (MCQ) dataset focused on elementary and middle school science. Similarly, OpenBookQA~\cite{openbookqa} is an MCQ dataset for elementary science facts. Additionally, several multilingual question-answering datasets have been introduced, such as T{\footnotesize Y}D{\footnotesize I}-QA, DuReader~\cite{he2017dureader}, and DRCD~\cite{shao2018drcd}.

CodeApex \cite{fu2023codeapex} evaluates programming comprehension through multiple-choice exam questions covering conceptual understanding, commonsense reasoning, and multi-hop reasoning, as well as code generation and code correction tasks. The authors tested 12 LLMs and found that GPT-4 exhibited the highest programming capabilities, with accuracies of 69\% in comprehension, 54\% in generation, and 66\% in correction tasks. They concluded that \textit{``Novice programmers perform similarly to GPT-4 in closed-book tests after learning, while human performance in open-book exams is significantly better than all LLMs.''} This trend was consistent across most models tested, including \texttt{Chinese-Alpaca-13B} and \texttt{InternLM-Chat-7B}. However, while GPT-4 outperformed humans in closed-book scenarios, humans were only slightly better in open-book exams.


In~\cite{liu_secqa_2023}, Z. Liu introduced a cybersecurity dataset designed to evaluate the capabilities of LLMs, but it is entirely based on a single material: \textit{"Computer Systems Security: Planning for Success."} It includes around two hundred questions. Notably, human comparison and validation were not part of the development process. 



\section{Methodology}
\label{sec:methodology}
The framework for creating the \texttt{CyberMetric} dataset is illustrated in Figure~\ref{fig:Framework}, and it comprises five crucial phases:  \ding{192} Data Collection, \ding{193} Question Generation, \ding{194} Question Post-processing, \ding{195} Question Validation, and \ding{196} Reference Dataset Creation Phase. 

\subsection{\ding{192} Data Collection}

The questions were generated by GPT-3.5 turbo using Retrieval-Augmented Generation (RAG)~\cite{NEURIPS2020_6b493230} from widely recognized cybersecurity documents, including open standards, NIST standards, research papers, publicly available books, RFCs, and other publications in the cybersecurity domain, totalling over 100,000 pages. As documents were in PDF format, we used \texttt{pdfminer}\footnote{\url{https://pypi.org/project/pdfminer/}} to extract the text, which was then segmented into chunks of 8000 token worth of text---well within the context window of GPT-3.5---to ensure it can be effectively processed.
During this stage, we excluded irrelevant sections such as tables of contents, prefaces, acknowledgements, pictures, references, and appendices. 

\subsection{\ding{193} Question Generation}
The chunks were fed to GPT-3.5-turbo to generate ten questions and corresponding multiple-choice answers for every 8,000 tokens. This approach aims to maintain a balanced representation of each publication. Creating 1,000 questions from a ten-page document might introduce redundancy to the dataset. 
The generated questions are reviewed by Falcon-180B, referred to as the \texttt{FALCON Content Review}, to identify grammatical and semantic errors and filter out irrelevant questions. By applying semantic analysis~\cite{zhang2023sentiment}, questions unrelated to cybersecurity were excluded. During this stage, human validators randomly examined the questions to assess the overall quality of the results, ensuring they are relevant to cybersecurity and written in clear, correct English.
GPT-3.5 generated a total of 11,000 questions, including an extra 10\% margin to allow for the removal of unnecessary or irrelevant questions. The \texttt{FALCON Content Review} module eliminated 1.7\% of the questions due to grammatical and semantic errors. Human experts then dedicated over 30 hours to eliminate an additional 2.3\% of the questions. The questions that remain form the \texttt{intermediate database}, containing grammatically correct and cybersecurity-focused questions.
However, at this stage, neither LLMs nor human experts have verified the accuracy of the solutions. Next, the 10,560 questions that were kept must undergo further solution validation to identify and remove any questions with multiple or incorrect solutions. 

\subsection{\ding{194} Question Post-processing}
First, we employ the cutting-edge T5-base model from Google \cite{google2021}, which is trained for grammar correction. 
This model identified 230 questions where the English could be improved. For instance, it flagged the sentence \textit{``Which of the following elements do not apply to privacy?''} for subject-verb agreement error. The correct form should be \textit{``Which of the following elements does not apply to privacy?''}.
Next, we have again utilized \texttt{FALCON-180B}, but this time instead of semantic checking, we aimed to eliminate non-contextual questions that cannot be understood without external reference. Questions generated during the creation phase, such as ``According to Figure 1'' or ``As seen in Table 6'', are nonsensical without relevant content.
After fixing and removing such issues, we have used GPT-4, GPT-4o, and Mistral7x8b to analyze all the remaining $10470$ questions, whether they think the provided answer is correct. 

\subsection{\ding{195} Test Phase}

We have reached a critical stage in our analysis where human experts must review all questions flagged by GPT-4, GPT-4o, and Mistral7x8b. Upon thoroughly examining the questions, we discovered that many were inaccurate or imprecise. These issues can be divided into four categories:

\begin{enumerate}
    \item \textbf{Multiple Correct Answers:} Some questions in the dataset had more than one correct answer provided.
    \item \textbf{Time Relevant Questions:} Since the questions and their answers are derived from sources up to ten years old, some information provided to GPT-3.5 may be outdated. For example, the answer to "Which web server dominates the market according to web surveys?" was Apache in 2015. However, as of 2024, the dominant web server is Nginx.

    \item \textbf{Incorrect Information in the Source:} To our surprise, some selected sources contained incorrect information despite being reputable and widely followed. We verified these inaccuracies with multiple field experts.
    
    \item \textbf{Missing References:} Some questions still reference material from the original document, such as ``As per table 2'' or ``According to Figure 1,'' making them unanswerable without the referred content. Despite prior efforts to eliminate such questions, we found new instances like ``In chapter 4'', or ``In the author's opinion.''

\end{enumerate}


After resolving the flagged issues, we further eliminated some redundant questions. This phase required more than 200 hours of expert human effort. The result forms the core of the \texttt{CyberMetric} dataset, containing exactly $10,000$ questions. Table~\ref{tab:distirbution} shows the final distribution of questions.

\begin{table}[ht]
\caption{CyberMetric Dataset: Questions Domain Distribution}
\scriptsize
\center
\begin{tabular}{cccc}\hline

\textbf{Domain} & \makecell{\textbf{Questions}\\\textbf{verified}} & \makecell{ \textbf{Number of}\\ \textbf{ Questions}} & \makecell{\textbf{Creation}\\ \textbf{Method}} \\\hline
\hline
\makecell{Penetration Testing /\\ Ethical Hacking} & \ding{52} & $1000$ & LLM \& Human \\ 
\hline
Cryptography & \ding{52} & $1500$ & LLM \& Human \\ 
\hline
\makecell{Network Security /\\ IoT Security}  & \ding{52} & $1000$ & LLM \& Human \\ 
\hline
\makecell{Information Security /\\ Information Governance}& \ding{52} & $1500$ & LLM \& Human \\ 
\hline
\makecell{ Compliance /\\ Disaster recovery}& \ding{52} & $1500$ & LLM \& Human \\ 
\hline
\makecell{Cloud Security  /\\Identity Management}& \ding{52} & $1500$ & LLM \& Human \\ 
\hline
\makecell{NIST guidelines /\\ RFC documents} & \ding{52} & $2000$ & LLM \& Human  \\ 
\hline

\rowcolor{gray!50}\textbf{CyberMetric}& \ding{52} & \textbf{$10000$} & LLM \& Human  \\ 
\hline

\end{tabular}
\\

\label{tab:distirbution}
\end{table}

\subsection{\ding{194} Reference Dataset Creation}
First, we create \textit{CyberMetric-80} for human participants, along with \textit{CyberMetric-500} where all answers are accurate, and verified by a panel of experts in various cybersecurity domains, each with a minimum of 10 years of experience and holding internationally recognized certifications like CISSP, CISM, OSCP, OSEP, and ISO 27001LA.

We recruited 30 survey participants through our social networks and contacted individuals from universities, research institutions, and Big4 consulting companies to volunteer. We have prepared a Google Forms survey containing inquiries related to gender, age, years of experience, and the highest level of education and requested volunteers, spanning from beginners to experts, to complete the questionnaire without using any additional help. 
Next, the 25 selected LLM models filled all the four \textit{CyberMetric} datasets. 
The comprehensive analysis of the accuracy of the 30 participants and the LLMs will be discussed in Section~\ref{sec:experiments}. 

Lastly, an automated statistical analysis is carried out for the entire set of questions. 
This is to confirm that all answers are consistent within the A, B, C, and D options. Additionally, we have taken steps to ensure that the answers are evenly distributed among the options.

\section{Experimental Results}
\label{sec:experiments}

The experiments are divided into three main parts: Comparing 25 LLMs cybersecurity knowledge in all the four \texttt{CyberMetric} datasets, assessing human vs LLM performance on \texttt{CyberMetric-80}, and determine the datasets accuracy.

\subsection{Assessing LLMs Performance}

The 25 LLM models were tested on an AWS ml.p4d.24xlarge instance with 8x NVIDIA Tesla A100 GPU cards and 382 GB of RAM, running Ubuntu 22.04.  The default settings were a temperature of 1.0, top\_p at 0.9, and top\_k at 50. 

The most accurate proprietary models were \texttt{GPT-4o} and \texttt{GPT-4-turbo} (see Table~\ref{tab:ranking}). The top-performing open-source models were \texttt{Mixtral-8x7B-Instruct} by Mistral AI and \texttt{Falcon-180B} by TII. Notably, the best performing small models with 7 billion parameters were \texttt{Mistral-7B-Instruct-v0.2} by Mistral AI, and \texttt{Gemma-1.1-7b-it} by Google. Further analysis of incorrect responses, weaknesses, and strengths will be discussed in Section~\ref{sec:discussion}. Table~\ref{tab:ranking} is ordered based on the column $2K\,\,Q$. Note, that due to the probabilistic nature of the models' outputs, the same model may vary by up to 3-4 percentage points in subsequent runs, even for the top-performing models.

\subsection{Human Performance on CyberMetric-80}

\begin{table}[t!]
\caption{Distribution of accuracy among participants.}

\center
\renewcommand{\arraystretch}{1.1}
\begin{tabular}{c|cccccccc}\hline
\multicolumn{6}{c}{\textbf{EXPERIENCED PARTICIPANTS}} \\ \hline
\textbf{\#} & \textbf{E} & \textbf{D}& \textbf{A}  & \textbf{G} & \textbf{R} \\\hline
\hline

\hline
P16  & 10+ & P.hD. & 35-50 & M &\cellcolor{green!80}88.75\% \\ \hline
P29  & 10+ & P.hD. & 35-50 & F &\cellcolor{green!75}87.50\% \\ \hline
P14  & 1-5 & MA/MSc & 35-50 & M & \cellcolor{green!70}87.50\% \\ \hline
P24  & 10+ & BA/BSc & 35-50 & M & \cellcolor{green!60}86.25\% \\ \hline
P26  & 1-5  & MA/MSc & 18-35 & M & \cellcolor{green!60}86.25\% \\ \hline
P13  & 10+ & P.hD. & 50+ & M & \cellcolor{green!50}83.75\% \\ \hline
P5  & 10+ & P.hD. & 35-50 & M & \cellcolor{green!40}82.50\% \\ \hline
P3  & 5-10 & MA/MSc & 35-50 & M & \cellcolor{green!40}82.50\% \\ \hline
P1  & 5-10 & MA/MSc & 18-35 & M & \cellcolor{green!30}76.25\% \\ \hline
P25  & 5-10 & BA/BSc & 35-50 & M & \cellcolor{green!25}75.00\% \\ \hline
P20  & 1-5 & Secondary   & 18-35 & M &\cellcolor{green!20}72.50\% \\ \hline
P30  & 5-10 & BA/BSc   & 18-35 & F &\cellcolor{green!23}71.25\% \\ \hline
P12  & 1-5 & MA/MSc & 18-35 & M & \cellcolor{green!23}71.25\% \\ \hline
P22  & 1-5 &  P.hD. & 18-35 & M &\cellcolor{green!15}70.00\% \\ \hline
P9  & 1-5 & BA/BSc & 18-35 & M & \cellcolor{green!15}70.00\% \\ \hline
P28  & 1-5 & Secondary  & 18-35 & M &\cellcolor{green!10}68.75\% \\ \hline
P2  & 1-5 & BA/BSc & 18-35 & M & \cellcolor{orange!15}58.75\% \\ \hline
P15  & 1-5 & MA/MSc & 18-35 & M & \cellcolor{orange!15}58.75\% \\ \hline
P21  & 1-5 & MA/MSc   & 18-35 & F &\cellcolor{orange!22}53.75\% \\ \hline
\hline
\multicolumn{5}{c}{Mean accuracy:}&$\approx$\textbf{72.24\%} \\ \hline
\hline
\multicolumn{6}{c}{\textbf{BEGINNERS}} \\ \hline
\textbf{\#} & \textbf{E} & \textbf{D}& \textbf{A}  & \textbf{G} & \textbf{R} \\\hline
\hline
P19  & 0 & BA/BSc  & 18-35 & M &\cellcolor{orange!8}63.75\% \\ \hline
P23  & 0 & MA/MSc  & 18-35 & M &\cellcolor{orange!14}61.25\% \\ \hline
P8  & 0 & BA/BSc & 35-50 & M & \cellcolor{orange!25}55.00\% \\ \hline
P27  & 0 & BA/BSc & 35-50 & M & \cellcolor{orange!25}55.00\% \\ \hline
P6  & 0 & MA/MSc & 35-50 & M & \cellcolor{orange!33}51.25\% \\ \hline
P18  & 0 & MA/MSc & 18-35 & F & \cellcolor{red!20}42.50\% \\ \hline
P11  & 0 & MA/MSc & 35-50 & F & \cellcolor{red!30}37.50\% \\ \hline
P4  & 0 & MA/MSc & 35-50 & F & \cellcolor{red!40}31.25\% \\ \hline
P7  & 0 &  BA/BSc &35-50 & F & \cellcolor{red!70}21.25\% \\ \hline
\hline
\multicolumn{5}{c}{Mean accuracy:}&$\approx$\textbf{46.58\%} \\ \hline
\hline

\multicolumn{6}{c}{\textbf{DISQUALIFIED PARTICIPANTS}} \\ \hline
\rowcolor{red!30} P10  & 0 & MA/MSc & 35-50 & F & 87.50\% \\ \hline
\rowcolor{red!30}   P17  & 0 & BA/BSc & 18-35 & F & 83.75\% \\ \hline

\hline

\end{tabular}
\\
\vspace{5pt}

Legend:\\ \textbf{E}: Experience  \textbf{D}: Degree, \textbf{A}: Age, \textbf{G}: Gender, \textbf{R}: Result
\label{tab:participants}
\end{table}

\begin{table*}[ht]
\centering
\caption{The 25 LLMs Performance on the CyberMetric Sorted by 2k Q Accuracy}
\label{tab:ranking}
\begin{tabular}{|c|ccc|cccc|}
\hline
\multirow{2}{*}{\textbf{LLM model}} & \multirow{2}{*}{\textbf{Company}} & \multirow{2}{*}{\textbf{Size}} & \multirow{2}{*}{\textbf{License}} & \multicolumn{4}{c|}{\textbf{Accuracy}} \\ \cline{5-8} 
                                    &                                   &                                &                                   & \multicolumn{1}{c|}{\textbf{80 Q}} & \multicolumn{1}{c|}{\textbf{500 Q}} & \multicolumn{1}{c|}{\textbf{2k Q}} & \textbf{10k Q} \\ \hline
\cellcolor{green!50} GPT-4o                         & \cellcolor{gray!20} OpenAI                            & \cellcolor{gray!20} N/A                            & \cellcolor{gray!20} Proprietary                       & \multicolumn{1}{c|}{96.25\%}                                & \multicolumn{1}{c|}{93.40\%}                                 & \multicolumn{1}{c|}{91.25\%}                                & 88.89\%                                 \\ \hline
\cellcolor{green!50} Mixtral-8x7B-Instruct               & Mistral AI                        & 45B                            & Apache 2.0                        & \multicolumn{1}{c|}{92.50\%}                                & \multicolumn{1}{c|}{91.80\%}                                 & \multicolumn{1}{c|}{91.10\%}                                & 87.00\%                                 \\ \hline
\cellcolor{green!50} GPT-4-turbo                         & \cellcolor{gray!20} OpenAI                            & \cellcolor{gray!20} N/A                            & \cellcolor{gray!20} Proprietary                       & \multicolumn{1}{c|}{96.25\%}                                & \multicolumn{1}{c|}{93.30\%}                                 & \multicolumn{1}{c|}{91.00\%}                                & 88.50\%                                 \\ \hline
\cellcolor{green!50} Falcon-180B-Chat                    & TII                               & 180B                           & Apache 2.0                        & \multicolumn{1}{c|}{90.00\%}                                & \multicolumn{1}{c|}{87.80\%}                                 & \multicolumn{1}{c|}{87.10\%}                                & 87.00\%                                 \\ \hline
\cellcolor{green!50} GPT-3.5-turbo                       & \cellcolor{gray!20} OpenAI                            & \cellcolor{gray!20} 175B                           & \cellcolor{gray!20} Proprietary                       & \multicolumn{1}{c|}{90.00\%}                                & \multicolumn{1}{c|}{87.30\%}                                 & \multicolumn{1}{c|}{88.10\%}                                & 80.30\%                                 \\ \hline
\cellcolor{green!50} GEMINI-pro 1.0                   & Google                            & 137B                           & Proprietary                       & \multicolumn{1}{c|}{90.00\%}                                & \multicolumn{1}{c|}{85.05\%}                                 & \multicolumn{1}{c|}{84.00\%}                                & 87.50\%                                 \\ \hline
\cellcolor{green!30} Mistral-7B-Instruct-v0.2            & \cellcolor{gray!20} Mistral AI                        & \cellcolor{gray!20} 7B                             & \cellcolor{gray!20} Apache 2.0                        & \multicolumn{1}{c|}{78.75\%}                                & \multicolumn{1}{c|}{78.40\%}                                 & \multicolumn{1}{c|}{76.40\%}                                & 74.82\%                                 \\ \hline
\cellcolor{green!30} Gemma-1.1-7b-it                     & Google                            & 7B                             & Open                              & \multicolumn{1}{c|}{82.50\%}                                & \multicolumn{1}{c|}{75.40\%}                                 & \multicolumn{1}{c|}{75.75\%}                                & 73.32\%                                 \\ \hline
\cellcolor{green!30} Meta-Llama-3-8B-Instruct     & \cellcolor{gray!20} Meta                        & \cellcolor{gray!20} 8B                            &    \cellcolor{gray!20} Open   & \multicolumn{1}{c|}{81.25\%}                                & \multicolumn{1}{c|}{76.20\%}                                 & \multicolumn{1}{c|}{73.05\%}                                & 71.25\%                                 \\ \hline
\cellcolor{green!15} Flan-T5-XXL                         & Google                            & 11B                            & Apache 2.0                        & \multicolumn{1}{c|}{81.94\%}                                & \multicolumn{1}{c|}{71.10\%}                                 & \multicolumn{1}{c|}{69.00\%}                                & 67.50\%                                 \\ \hline
\cellcolor{green!15} Llama 2-70B                         & \cellcolor{gray!20} Meta                              & \cellcolor{gray!20}70B                            & \cellcolor{gray!20} Apache 2.0                        & \multicolumn{1}{c|}{75.00\%}                                & \multicolumn{1}{c|}{73.40\%}                                 & \multicolumn{1}{c|}{71.60\%}                                & 66.10\%                                 \\ \hline
\cellcolor{green!15} Zephyr-7B-beta                      & HuggingFace                       & 7B                             & MIT                               & \multicolumn{1}{c|}{80.94\%}                                & \multicolumn{1}{c|}{76.40\%}                                 & \multicolumn{1}{c|}{72.50\%}                                & 65.00\%                                 \\ \hline
\cellcolor{green!15} Qwen1.5-MoE-A2.7B                   & \cellcolor{gray!20} Qwen                              & \cellcolor{gray!20} 2.7B                           & \cellcolor{gray!20} Open                              & \multicolumn{1}{c|}{62.50\%}                                & \multicolumn{1}{c|}{64.60\%}                                 & \multicolumn{1}{c|}{61.65\%}                                & 60.73\%                                 \\ \hline
\cellcolor{green!15} Qwen1.5-7B                          & Qwen                              & 7B                             & Open                              & \multicolumn{1}{c|}{73.75\%}                                & \multicolumn{1}{c|}{60.60\%}                                 & \multicolumn{1}{c|}{61.35\%}                                & 59.79\%                                 \\ \hline
\cellcolor{orange!30} Qwen-7B                             & \cellcolor{gray!20} Qwen                              & \cellcolor{gray!20} 7B                             & \cellcolor{gray!20} Open                              & \multicolumn{1}{c|}{43.75\%}                                & \multicolumn{1}{c|}{58.00\%}                                 & \multicolumn{1}{c|}{55.75\%}                                & 54.09\%                                 \\ \hline
\cellcolor{orange!30} Phi-2                               & Microsoft                         & 2.7B                           & MIT                               & \multicolumn{1}{c|}{53.75\%}                                & \multicolumn{1}{c|}{48.00\%}                                 & \multicolumn{1}{c|}{52.90\%}                                & 52.13\%                                 \\ \hline
\cellcolor{orange!30} Llama3-ChatQA-1.5-8B      & \cellcolor{gray!20} Nvidia               & \cellcolor{gray!20} 8B & \cellcolor{gray!20} Open         & \multicolumn{1}{c|}{ 53.75\%}                                & \multicolumn{1}{c|}{ 52.80\%}                                 & \multicolumn{1}{c|}{49.45\%}                                &  49.64\%                                 \\ \hline
\cellcolor{orange!30} DeciLM-7B                           & Deci                              & 7B                             & Apache 2.0                        & \multicolumn{1}{c|}{52.50\%}                                & \multicolumn{1}{c|}{47.20\%}                                 & \multicolumn{1}{c|}{50.44\%}                                & 50.75\%                                 \\ \hline
\cellcolor{red!25} Qwen1.5-4B                          & \cellcolor{gray!20} Qwen                              & \cellcolor{gray!20} 4B                             & \cellcolor{gray!20} Open                              & \multicolumn{1}{c|}{36.25\%}                                & \multicolumn{1}{c|}{41.20\%}                                 & \multicolumn{1}{c|}{40.50\%}                                & 40.29\%                                 \\ \hline
\cellcolor{red!25} Genstruct-7B   & NousResearch       & 7B                             & Apache 2.0     & \multicolumn{1}{c|}{38.75\%}                                & \multicolumn{1}{c|}{40.60\%}                                 & \multicolumn{1}{c|}{37.55\%}                                & 36.93\%                                 \\ \hline
\cellcolor{red!25} Meta-Llama-3-8B                     & \cellcolor{gray!20} Meta                              & \cellcolor{gray!20} 8B                             & \cellcolor{gray!20} Open                              & \multicolumn{1}{c|}{38.75\%}                                & \multicolumn{1}{c|}{35.80\%}                                 & \multicolumn{1}{c|}{37.00\%}                                & 36.00\%                                 \\ \hline
\cellcolor{red!25} Gemma-7b                            & Google                            & 7B                             & Open                              & \multicolumn{1}{c|}{42.50\%}                                & \multicolumn{1}{c|}{37.20\%}                                 & \multicolumn{1}{c|}{36.00\%}                                & 34.28\%                                 \\ \hline
\cellcolor{red!50} Dolly V2 12b BF16                   & \cellcolor{gray!20} Databricks                        & \cellcolor{gray!20} 12B                            & \cellcolor{gray!20} MIT                               & \multicolumn{1}{c|}{33.75\%}                                & \multicolumn{1}{c|}{30.00\%}                                 & \multicolumn{1}{c|}{28.75\%}                                & 27.00\%                                 \\ \hline
\cellcolor{red!50} Gemma-2b                            & Google                            & 2B                             & Open                              & \multicolumn{1}{c|}{25.00\%}                                & \multicolumn{1}{c|}{23.20\%}                                 & \multicolumn{1}{c|}{18.20\%}                                & 19.18\%                                 \\ \hline
\cellcolor{red!90} Phi-3-mini-4k-instruct              & \cellcolor{gray!20} Microsoft                         & \cellcolor{gray!20} 3.8B                           & \cellcolor{gray!20} MIT                               & \multicolumn{1}{c|}{5.00\%}                                 & \multicolumn{1}{c|}{5.00\%}                                  & \multicolumn{1}{c|}{4.41\%}                                 & 4.80\%                                  \\ \hline
\end{tabular}
\end{table*}

Altogether, $30$ participants completed the \texttt{CyberMetric-80} questionnaire which takes an expert around 40 to 60 minutes. They were all instructed to take a closed-book exam. We identified two participants due to their unusually high accuracy levels despite lacking cybersecurity experience. They used GPT-3.5, thus their accuracy and 92\% of their incorrect responses matched those produced by GPT-3.5. These participants confirmed our suspicion and were excluded from the analysis.

\begin{figure}[b] 
\centering
\includegraphics[width=0.5\textwidth]{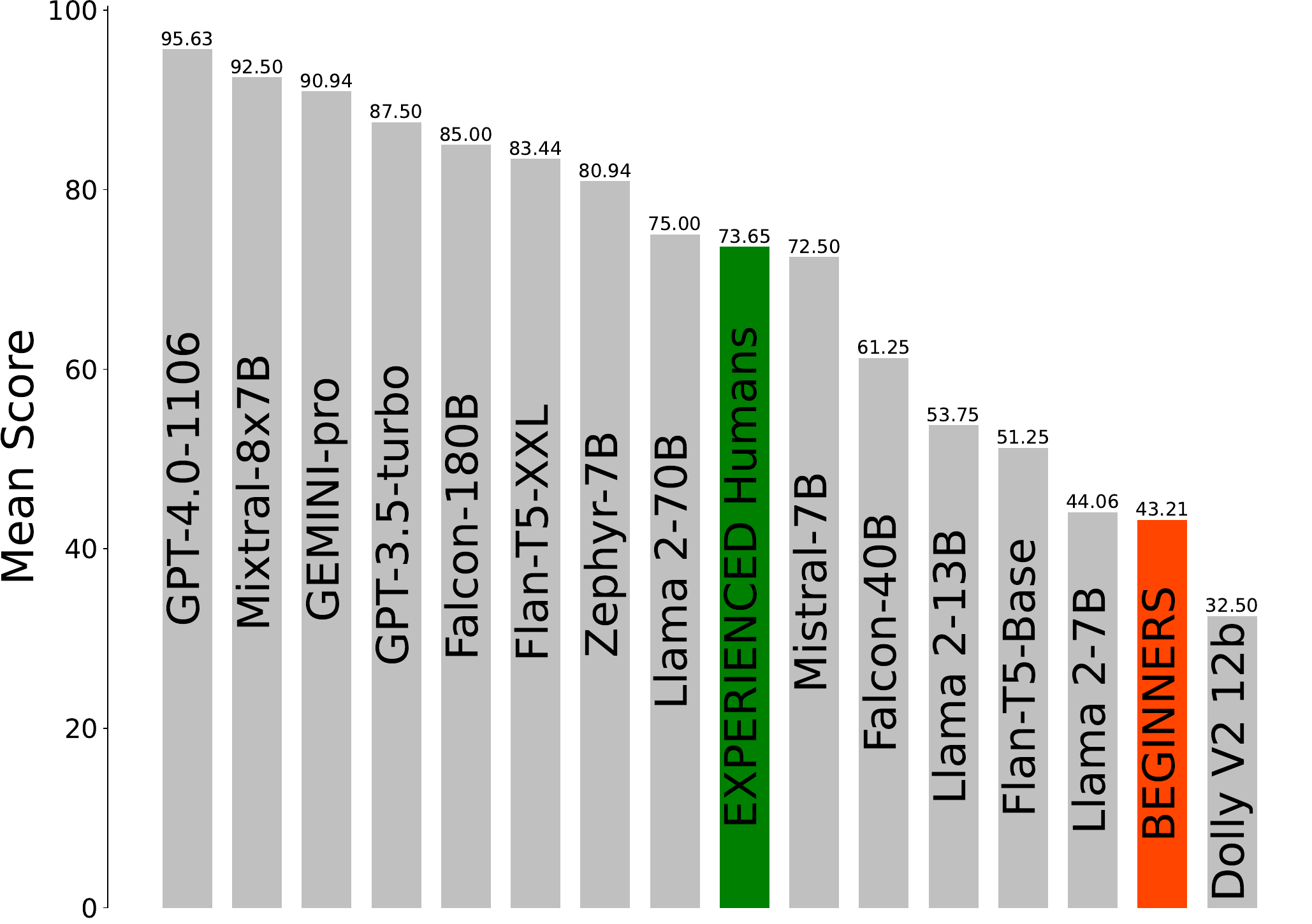}
\caption{Comparing Human vs LLM performance on CyberMetric-80}
\label{fig:cucc}
\end{figure}

The \texttt{CyberMetric} dataset covers a diverse range of topics, making it challenging for human experts to recall everything from memory in a closed-book scenario, without prior preparation. The highest score is $71/80\approx 88.75\%$ by an individual holding a PhD and certifications in CISSP, OSCP, and ISO27001LA. The median score among the $30$ participants is $56$, while the mean accuracy is $53.83$. The mean accuracy for experienced participants (with at least 1-5 years of cybersecurity experience) is $72.24$\%. Beginners with no prior cybersecurity experience achieved scores ranging from 21.25\% to 63.75\%. Table \ref{tab:participants} shows the results of all participants.
Figure~\ref{fig:cucc} compares humans against LLMs on CyberMetric 80. The percentages for the LLMs are derived another test run, leading to slight variations from the results shown in Table~\ref{tab:ranking}.

\subsection{Dataset Accuracy}

As noted, \texttt{CyberMetric-80} and \texttt{CyberMetric-500} have all questions and answers fully validated by human experts and can serve as reference points. If there is a significant difference in the LLM's accuracy between \texttt{CyberMetric-10000} and the fully validated \texttt{CyberMetric-80} and \texttt{CyberMetric-500} datasets (e.g., 30\%), it would suggest that the larger dataset lacks precision and some solutions might still be incorrect. As shown in TABLE~\ref{tab:ranking}, this is not the case. While most models has less accuracy on \texttt{CyberMetric-10000} compared to the smaller subsets, it is also due to the fact that proportionally there are more questions on new recommendations, guidelines, and recent research, which many models are unfamiliar with, also contributing to the drop in accuracy. We estimate that 2-3\% of the questions in \texttt{CyberMetric-10000} still have issues outlined in Section \ref{sec:methodology}-D. We encourage readers and researchers to report any inadequate or questionable answers by opening an issue on the dataset's GitHub repository.

\section{Discussion - Human vs Machine}
\label{sec:discussion}

On average, expert humans, with a mean accuracy of 72\%, perform at a level comparable to \texttt{Llama 2-70B}. The top human performers are close to Falcon-180B, GPT-3.5, and GEMINI-pro 1.0. GPT-4o, Mixtral-8x7B-Instruct, and GPT-4-turbo significantly outperform any human beings on the \texttt{CyberMetric-80} dataset.

\subsection{Most Difficult Questions for Humans on CyberMetric-80}
The analysis revealed the questions that posed the greatest challenge to participants. Table~\ref{tab:most_challenge_for_humans} highlights the top 5 questions where all participant responses were at or below a 25\% success rate.
Questions 41 and 66 posed similar challenges to both humans and LLMs. This similarity can be attributed to these questions being relatively new recommendations from NIST and BSI. 



\begin{table}[b!]
\centering
\caption{Most challenging questions for Humans. \\(With multiple choice answers provided)}
\label{tab:most_challenge_for_humans}
\renewcommand{\arraystretch}{1.2} 
\begin{tabular}{|c|p{0.4\textwidth}|c|}
\hline
\rowcolor[HTML]{EFEFEF} 
\textbf{No.} & \textbf{Questions} \\ \hline
\textbf{Q41} & What is the essential requirement for the security of the Discrete Logarithm Integrated Encryption Scheme (DLIES) to be maintained?  \\ \hline
\rowcolor[HTML]{EFEFEF} 
\textbf{Q42} & Which document provides the recommendation for elliptic curve cryptography?  \\ \hline   
\textbf{Q45} & What is the purpose of implementing monitoring systems?  \\ \hline
\textbf{Q61} & What is the primary goal of an Information Security Governance strategy? \\ \hline
\textbf{Q66} & According to NIST guidelines, what is a recommended practice for creating a strong password?  \\ \hline

\end{tabular}
\end{table}

\subsection{Most Difficult Questions for LLMs}

Which domain poses the greatest challenges for LLMs? This question is crucial for meaningful comparisons and facilitates future research. Do humans encounter the same difficulties, or are these challenges distinct to LLMs? Here, we present fascinating findings, highlighting the two most problematic domain for LLMs in answering questions.

\subsubsection{Difficulty 1: Cutting Edge Research} 

LLMs often struggle with accurately responding to questions based on the latest research, especially when their training data is outdated. For instance, consider the question: ``What is the recommended interval for changing passwords?'' Historically, many guidelines mandated a 30-day password change interval. Consequently, models trained on extensive data typically endorse this interval. However, the latest NIST SP 800-63B Section 5.1.1.2 paragraph 9 advises against periodic changes, recommending password changes only if there is evidence of compromise. The official explanation, while simple, is not entirely straightforward as it challenges the established research paradigm of password behaviours that has prevailed for the past 30 years. Recent research indicates that users tend to select weaker passwords when they know they must change them soon. Since most documents do not align with these updated guidelines, it is anticipated that most LLMs will recommend users to change their passwords. Only a few models, like GPT-4o and GPT-4, can accurately respond to these factual questions, possibly due to internet access or recent training data. 

Another question that advanced language models and humans often find challenging concerns the latest Federal Office for Information Security (BSI) recommendations. In January 2023, the BSI released the \textit{"Cryptographic Mechanisms: Recommendations and Key Lengths (BSI TR-02102-1)"} guideline. This guideline specifies that for the Discrete Logarithm Integrated Encryption Scheme (DLIES), the prime number $p$ should be no less than 3000 bits, and the prime number \( q \) should be a minimum of 250 bits. The choice of $3000$ bits for $p$, which is not a power of two, and the requirement for $q$  to be $256$ bits results in what appears to be an unbalanced prime scheme. Even expert security analysts often find this question challenging, and many language models struggle to provide the correct answer when dealing with such a question.

\subsubsection{Difficulty 2: Complex Computations}
LLMs often face challenges in scenarios requiring precise calculations due to their lack of access to RAG or external tools. For instance, a question like \textit{"What is the result of the bitwise XOR operation between 110101 and 101010 in binary?"} can be difficult. While a model like GPT-4o can handle this easily with an embedded Python tool, models such as \texttt{Mixtral-8x7B-Instruct} or \texttt{GPT-3.5} may fail without such tools.

Similarly, a straightforward question like \textit{``What is the CIDR notation equivalent for the subnet mask 255.255.248.0?''} should be easily answerable. However, many models struggle to respond correctly without external assistance. Complex models often produce incorrect answers for precise calculations, highlighting the need for external tools to solve intricate mathematical or reasoning problems.

\section{Limitations and Ethical Considerations}
\label{sec:limit}

\subsection{Limitations and Threats to Validity}
For \texttt{CyberMetric-10000}, most of the questions have been validated by human experts. However, there remains the possibility of incorrect validations or the presence of irrelevant questions. Updates and corrections will be announced on the project's website: \url{https://github.com/CyberMetric}.

\subsection{Ethical Considerations}
The documents used in this study are publicly accessible via internet searches. \texttt{CyberMetric} incorporates a diverse range of standard and open-access documents from the security field, including standards, research papers, NIST special publications, BSI guidelines, and RFC documents. During the human validation phase, the authors took all necessary measures to eliminate non-publicly available content from the questions or validate the source when RAG is used to generate the questions.

\section{Conclusion}
\label{sec:conclusion}
In this research, we introduced the \texttt{CyberMetric} dataset to evaluate the broad cybersecurity knowledge of LLMs. Our study focused on answering two key research questions:

\begin{itemize}

    \item {\textbf{RQ1}:} Has machine intelligence already surpassed humans in answering questions across the entire breadth of cybersecurity knowledge in a closed-book test?
    
    \textbf{Answer}: Yes, in our study, GPT-4o outperformed all human experts on the \texttt{CyberMetric-80} test, indicating that machine intelligence has surpassed human performance in knowledge based cybersecurity questions in a closed-book scenario. Expert humans, with a mean accuracy of 72\%, were on par with models like \texttt{Llama 2-70B}. While top human performers nearly matched the highest-performing LLMs. Beginners lagged significantly, being outperformed by 18 of the 25 models tested.
       \item {\textbf{RQ2}:} Which currently available model achieves the highest accuracy in answering questions across diverse cybersecurity domains?
    
    \textbf{Answer}: GPT-4o and GPT-4 were identified as the top proprietary models in terms of accuracy. Among open-source models, Mixtral-8x7B-Instruct and Falcon-180B performed best, achieving identical scores on \texttt{CyberMetric-10000}.
\end{itemize}

Most models still experience limitations in complex calculations and reasoning. Note that this dataset is not suitable for drawing general conclusions about machine versus human intelligence, intuition, and problem-solving skills in general, as \texttt{CyberMetric} exclusively focuses on cybersecurity questions and answering knowledge. Nevertheless, we are witnessing a pivotal era where machines increasingly excel beyond human abilities in many aspects.


\section*{Acknowledgment}
We extend our heartfelt gratitude to the volunteers in the \texttt{CyberMetric-80} survey. Their dedication in completing this intricate and time-consuming 80-question test is immensely appreciated and vital for the success of this research. In an era where even support for a brief 5-minute survey is challenging to secure, the commitment shown by our participants is highly appreciated.



%
\bibliographystyle{plain}
\bibliography{references}

\end{document}